\PassOptionsToPackage{utf8}{inputenc}
\documentclass{bioinfo}
\copyrightyear{2018} \pubyear{2018}

\usepackage{tabularx}
\usepackage{array}
\usepackage{graphicx} 
\usepackage{multirow}
\usepackage{hhline,booktabs,amsmath}
\usepackage{makecell}
\usepackage{url}
\usepackage{todonotes}
\usepackage{subcaption}
\usepackage{amsmath}
\usepackage[linesnumbered,ruled]{algorithm2e}
\renewcommand{\cite}{\citep}
\newcommand\longvar[1]{\mathchardef\UrlBreakPenalty=100
\mathchardef\UrlBigBreakPenalty=100\url{#1}}

\access{Advance Access Publication Date: Day Month Year}
\appnotes{Manuscript Category}

\begin{document}
\firstpage{1}

\subtitle{Subject Section}

\title[OPA2Vec]{OPA2Vec: combining formal and informal
  content of biomedical ontologies to improve similarity-based
  prediction}

\author[Smaili \textit{et~al}.]{Fatima Zohra Smaili\,$^{\text{\sfb
      1}}$, Xin Gao\,$^{\text{\sfb 1,}*}$ and Robert
  Hoehndorf\,$^{\text{\sfb 1,}*}$}

\address{$^{\text{\sf 1}}$King Abdullah University of Science and
  Technology (KAUST), Computational Bioscience Research Center (CBRC),
  Computer, Electrical \& Mathematical Sciences and Engineering
  (CEMSE) Division, Thuwal 23955, Saudi Arabia.}

\corresp{$^\ast$To whom correspondence should be addressed.}

\history{Received on XXXXX; revised on XXXXX; accepted on XXXXX}

\editor{Associate Editor: XXXXXXX}

\abstract{\textbf{Motivation:} Ontologies are widely used in biology
  for data annotation, integration, and analysis. In addition to
  formally structured axioms, ontologies contain meta-data in the form
  of annotation axioms which provide valuable pieces of information
  that characterize ontology classes. Annotations commonly used in
  ontologies include class labels, descriptions, or synonyms. Despite
  being a rich source of semantic information, the ontology meta-data
  are generally unexploited by ontology-based analysis methods such as
  semantic similarity measures. \\
  \textbf{Results:} We propose a novel method, OPA2Vec, to generate
  vector representations of biological entities in ontologies by
  combining formal ontology axioms and annotation axioms from the
  ontology meta-data. We apply a Word2Vec model that has been
  pre-trained on PubMed abstracts to produce feature vectors from our
  collected data.  We validate our method in two different ways:
  first, we use the obtained vector representations of proteins as a
  similarity measure to predict protein-protein interaction (PPI) on
  two different datasets. Second, we evaluate our method on predicting
  gene-disease associations based on phenotype similarity by
  generating vector representations of genes and diseases using a
  phenotype ontology, and applying the obtained vectors to predict
  gene-disease associations.  These two experiments are just an
  illustration of the possible applications of our method. OPA2Vec can
  be used to produce vector representations of any
  biomedical entity given any type of biomedical ontology.\\
  \textbf{Availability:}
  \url{https://github.com/bio-ontology-research-group/opa2vec}\\
  \textbf{Contact:}
  \href{robert.hoehndorf@kaust.edu.sa}{robert.hoehndorf@kaust.edu.sa
    and xin.gao@kaust.edu.sa.}}

\maketitle

\section{Introduction}
Biological knowledge is widely-spread across different types of
resources. Biomedical ontologies have been highly successful in
providing the means to integrate data across multiple disparate
sources by providing an explicit and shared specification of a
conceptualization of a domain \cite{Gruber1995}.  Notably, ontologies
provide a structured and formal representation of biological knowledge
through logical axioms \cite{Hoehndorf2015role}, and ontologies are
therefore widely used to capture information that is extracted from
literature by biocurators \cite{biontologies}. However, ontologies do
not only include a formal, logic-based structure but also include many
pieces of meta-data that are primarily intended for human use, such as
labels, descriptions, or synonyms \cite{Smith2007}. 

Due to the pervasiveness of ontologies in the life sciences, many
applications have been built that exploit various aspects of
ontologies for data analysis and to construct predictive models.  For
example, a wide selection of semantic similarity measures have been
developed to exploit information in ontologies \cite{resnik,
  lin1998information, jiang1997semantic, wupalmer, leacock1998,
  li2003approach, al2006cluster}, and semantic similarity measures
have successfully been applied to the prediction of protein-protein
interactions \cite{Couto2009}, gene-disease associations
\cite{Koehler2009}, or drug targets \cite{Hoehndorf2013drugs}.

Recently, a set of methods have been developed that can characterize
heterogeneous graphs through ``embeddings'', i.e., methods to generate
knowledge graph embedding methods \cite{transe, Nickel2016,
  ristoski2016rdf2vec}. These methods are used to produce feature
vectors of knowledge embedded in heterogeneous graphs (i.e., knowledge
graphs), and they have already been applied successfully in the
biomedical domain \cite{alsharani17}. However, ontologies, in
particular those in the biomedical domain, cannot easily be
represented as graphs \cite{Rodriguez-Garcia2018} but rather
constitute logical theories that are best represented as sets of
axioms \cite{dlhandbook}.

Recently we developed Onto2Vec, a method that generates feature
vectors from the formal logical content of ontologies
\cite{smaili2018onto2vec}, and we could demonstrate that Onto2Vec can
outperform existing semantic similarity measures. Here, we extend
Onto2Vec to OPA2Vec (Ontologies Plus Annotations to Vectors) to
jointly produce vector representations of entities in biomedical
ontologies based on both the semantic content of ontologies (i.e., the
logical axioms) and the meta-data contained in ontologies as
annotation axioms.  We combine multiple types of information contained
in biomedical ontologies, including asserted and inferred logical
axioms, datatype properties, and annotation axioms to generate a
corpus that consists of both formal statements, natural language
statements, and annotations that relate entities to literals.
We then apply a Word2Vec model 
to generate vector representations for any entity named in the
ontology. Using transfer learning, we apply a pre-trained Word2Vec
model in OPA2Vec to significantly improve the performance in encoding
natural language phrases and statements.

We evaluate OPA2Vec using two different ontologies and applications:
first, we use the Gene Ontology (GO) \cite{GO} to produce vector
representations of yeast and human proteins and determine their
functional similarity and predict interactions between them; second,
we evaluate our method on the PhenomeNET ontology
\cite{hoehndorf2011phenomenet,rodriguez2017integrating} to infer
vector representations of genes and diseases and use them to predict
gene-disease associations. 
We demonstrate that OPA2Vec can produce task-specific and trainable
representations of biological entities that significantly outperform
both Onto2Vec and traditional semantic similarity measures in
predicting protein-protein interactions and gene-disease
associations. OPA2Vec is a generic method which can be applied to any
ontology formalized in the Web Ontology Language (OWL) \cite{Grau2008,
  owl2}. 


\section{Results}

\subsection{Encoding ontologies plus annotations as vectors}
OPA2Vec is an algorithm that uses asserted and inferred logical axioms
in ontologies, combines them with annotation axioms (i.e., meta-data
associated with entities or axioms in ontologies) and produces dense
vector representations of all entities named in an ontology, or
entities associated with classes in an ontology.  Ontologies
formalized in the Web Ontology Language (OWL) \cite{Grau2008} are
based on a Description Logic \cite{dlhandbook}. In Description Logics,
an ontology is described as the combination of a TBox and an ABox
\cite{HorrocksKS06}. The TBox is a set of axioms that formally
characterize classes (e.g., {\tt behavior SubClassOf: 'biological
  process'}), while the ABox contains a set of axioms that
characterize instances (e.g., {\tt P0AAF6 instanceOf: hasFunction some
  behavior}). The TBox and ABox together are used by the Onto2Vec
method \cite{smaili2018onto2vec} to generate dense vector
representations; to achieve this goal, Onto2Vec treats asserted or
inferred axioms as sentences which form a corpus, and vectors are
generated using Word2Vec \cite{word2vec1, word2vec2}.

In addition to the TBox and ABox (i.e., to the formal, logical
content), ontologies contain a large amount of meta-data in the form
of annotation axioms \cite{Hoehndorf2015role, Smith2007}. Ontology
meta-data consist of the set of non-logical annotation axioms that
describe different aspects of ontology classes, relations, or
instances. For example, most ontologies associate entities with a
label, a natural language description, several synonyms, etc. While
such meta-data are distinct from the formal content of an ontology and
therefore not exploited by methods such as Onto2Vec, they nevertheless
provide valuable information about ontology classes, relations, and
instances.

OPA2Vec (Ontologies Plus Annotations to Vectors) extends Onto2Vec to
combine both the formal content of ontologies and the meta-data
expressed as annotation axioms to generate feature vectors for any
named entity in an ontology; the vectors encode for both the formal
and informal content that characterize and constrain the entities in
an ontology.

Our algorithm generates sentences from OWL annotation axioms to form a
corpus. For example, from the assertion that an OWL class $C$ has a
label $L$ (using the {\tt rdfs:label} in the OWL annotation axiom) we
generate the sentence {\tt C rdfs:label L} (using the complete IRIs
for $C$, $L$, and {\tt rdfs:label}). If $C$ has an annotation axioms
relating it to multiple words or sentences using the annotation
property $r_a$, such as when providing a textual definition or
description for a class (e.g., using the Dublin Core {\tt description}
property), we generate a single sentence in which $C$ is related using
$r_a$ to the complete value of the annotation property (i.e., we
ignore sentence or paragraph delimiters). Some annotation properties
do not relate entities to strings, but, for example, to dates,
numbers, or other literals. For example, an ontology may contain
information about the creation date of a class or axiom; we also
generate sentences from these annotation axioms and render the value
of the annotation property as a string.

In OPA2Vec, we combine the corpus generated from the meta-data (i.e.,
annotation axioms) and the inferred and asserted logical axioms (using
the Onto2Vec algorithm).  We then apply a Word2Vec skipgram model on
the combined corpus to generate vector representations of all entities
in the ontology (for technical details, see Section 4.3).

Natural language words used in annotation properties have a
pre-defined meaning which cannot easily be derived from their use
within an ontology alone. Therefore, we use transfer learning in
OPA2Vec to assign a semantics to natural language words based on their
use in a large corpus of biomedical text.  In particular, we pre-train
a Word2Vec model on all PubMed abstracts so that natural language
words are assigned a semantics (and vector representation) based on
their use in biomedical literature (see Section 4.2). The vocabulary
in biomedical literature overlaps with the values of annotation
properties (i.e., the natural language words used to describe entities
in ontologies) but is disjoint with the vocabulary generated by
Onto2Vec (i.e., the IRIs that make up the classes, relations, and
instances in an ontology). In OPA2Vec, we therefore update the
pre-trained Word2Vec model to generate vectors for the entities in the
ontology, and to update the representations of words that overlap
between PubMed abstracts and the ontology annotations.

\begin{figure*}[ht]
  \centering
  \includegraphics[width=.9\textwidth]{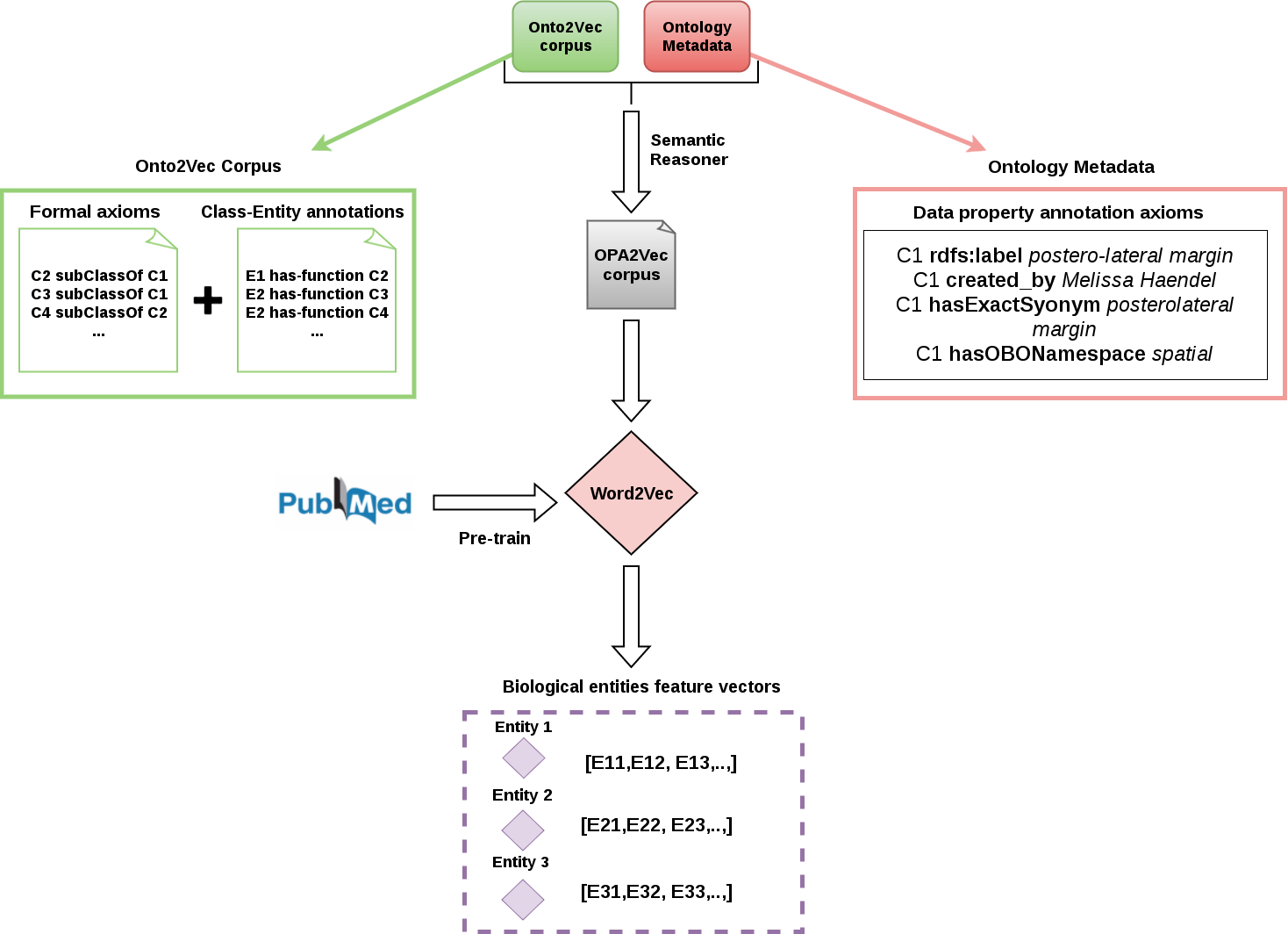}
  \caption{The detailed workflow of the feature vector generation pipeline of OPA2Vec}
  \label{fig:workflow}
\end{figure*}

Figure \ref{fig:workflow} illustrates the OPA2Vec algorithm. The input
of the algorithm is an ontology $O$ in the OWL format as well as a set $A$
of instances and their associations with classes in the ontology
(formulated as the OWL axioms). The output of OPA2Vec is a vector
representation for each entity in $O$ and $A$ that encodes for the
logical axioms and meta-data in $O$ and $A$.

		


\subsection{OPA2Vec performance in predicting interactions between
  proteins}



Ontologies are widely used to analyze biological and biomedical
datasets \cite{Hoehndorf2015role}, and one of the main applications of
ontologies is the computation of semantic similarity \cite{Couto2009}.
As OPA2Vec combines logical axioms and annotation axioms into single
vector representations, we expect that we can obtain more accurate
feature vectors for biological entities than using the ontology
structure alone, and that we can use this to improve the computation of
semantic similarity.

\begin{figure*}[ht]
	\centering
	\begin{subfigure}{.5\textwidth}
		\centering
		\includegraphics[width=\linewidth, trim=4 4 4 4,clip]{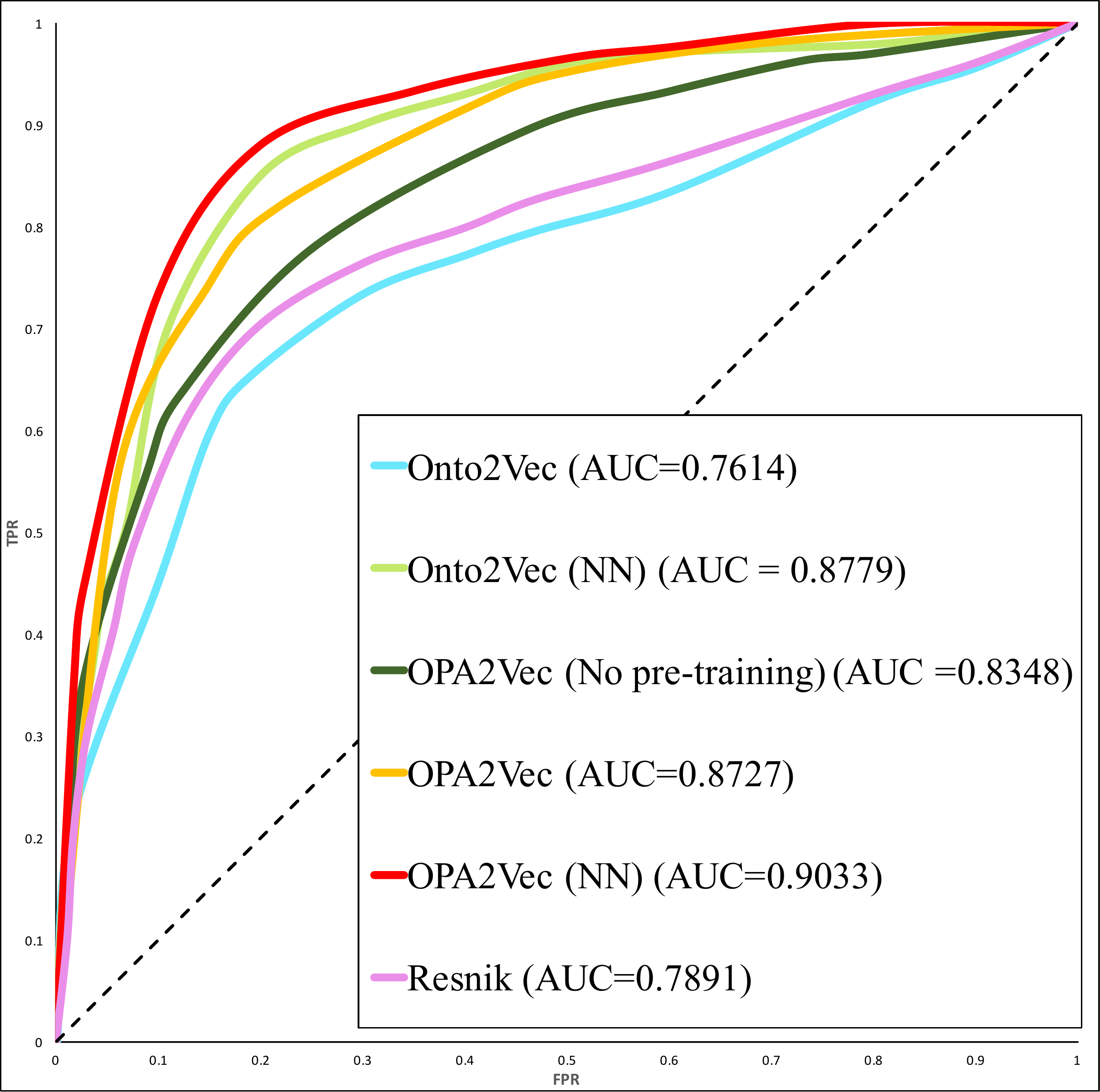}
		\caption{Human}
		\label{fig:sub1}
	\end{subfigure}%
	\begin{subfigure}{.5\textwidth}
		\centering
		\includegraphics[width=\linewidth,trim=4 4 4 4,clip,]{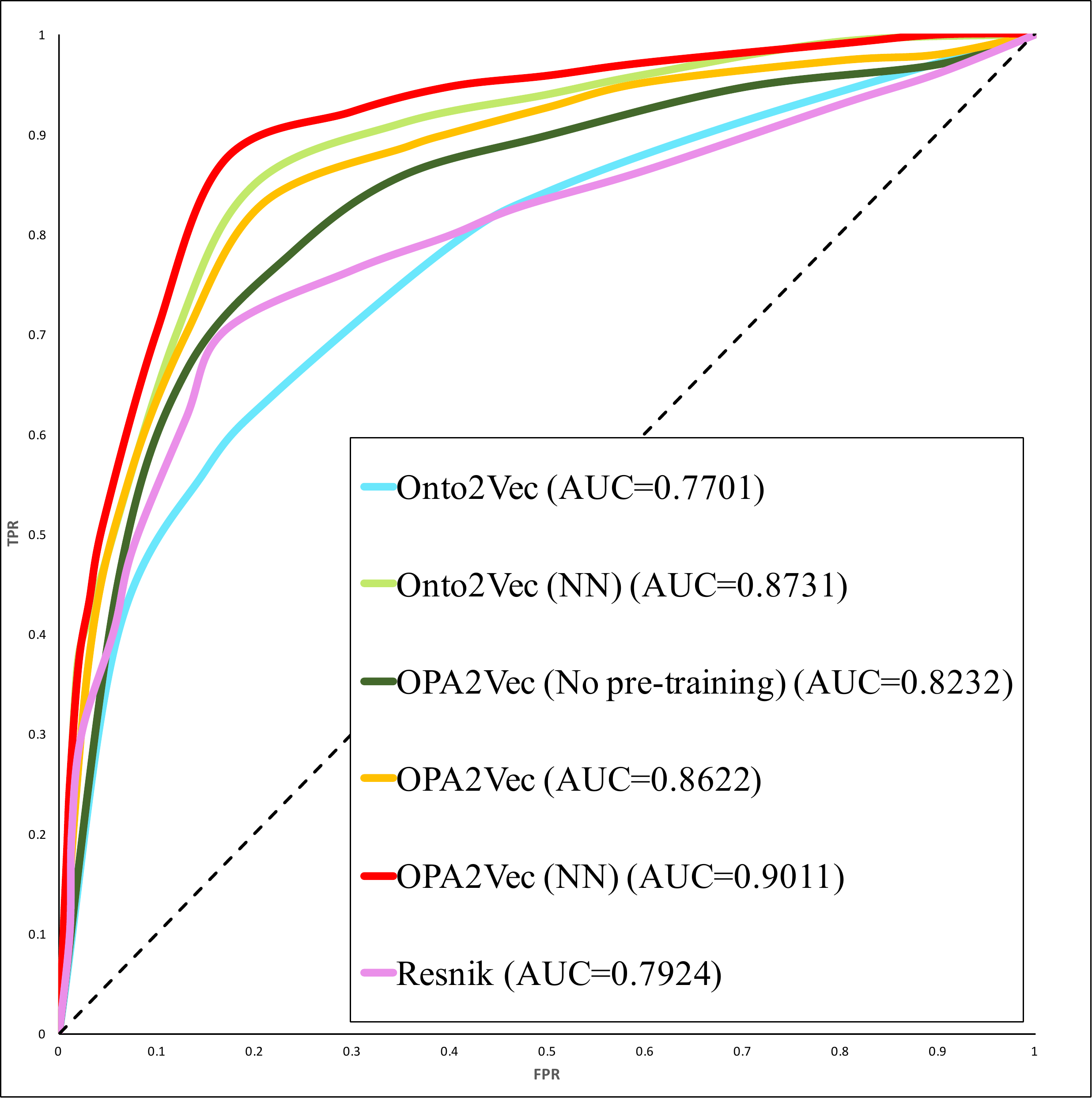}
		\caption{Yeast}
		\label{fig:sub2}
	\end{subfigure}
	\caption{\label{fig:ppiprediction}ROC curves and AUC values of
          different methods for PPI prediction for yeast and
          human. {\tt Onto2Vec} uses formal ontology axioms and
          compares vectors through cosine similarity; {\tt
            Onto2Vec(NN)} uses a neural network to compare vectors;
          {\tt OPA2Vec} is our method and uses formal ontology axioms,
          entity-class associations and annotation properties from the
          ontology meta-data (labels, description, synonyms,
          created\_by) with a Word2Vec model pre-trained on PubMed,
          and compares vectors through cosine similarity; {\tt
            OPA2Vec(No pre-training)} uses same strategy as OPA2Vec
          but without a pre-trained Word2Vec model; {\tt OPA2Vec(NN)}
          is {\tt OPA2Vec} and uses a neural network to determine
          similarity between two protein vectors; {\tt Resnik} is a
          semantic similarity measure.}
\end{figure*}

\begin{figure}[ht!]
	\centering
	\includegraphics[width=.47\textwidth]{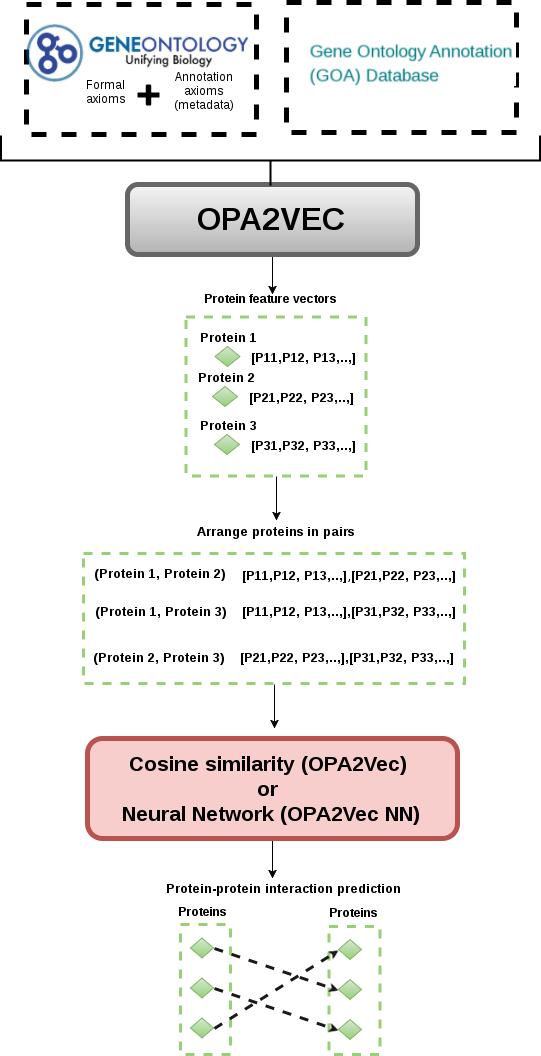}
	\caption{Workflow for protein-protein interaction (PPI) prediction using OPA2Vec.}
	\label{fig:workflowppi}
\end{figure}

To evaluate our hypothesis and demonstrate the potential of using
OPA2Vec, we used the GO ontology as a case study (see Section 4.1). We
generated a knowledge base using GO, and added either human proteins
or yeast proteins as instances. We related each protein to its
functions by asserting that a protein $P$ with function $F$ is an
instance of the class {\tt has-function some F}. We applied OPA2Vec on
these two knowledge bases (one including human proteins and the other
yeast proteins) and generated vector representations for each protein
and ontology class.  We then used these vector representations to
predict interactions between proteins as characterized in the STRING
database \cite{string} by calculating the cosine similarity between
each pair of protein vectors and using the obtained value as a
prediction score for whether two proteins interact or not. To further
improve our prediction performance, we used a neural network model to
learn a similarity measure between two feature vectors that is
predictive of protein-protein interactions \cite{smaili2018onto2vec}.
Figure \ref{fig:ppiprediction} shows the ROC curves and AUC values
obtained for OPA2Vec, and the comparison results against Onto2Vec and
Resnik's semantic similarity measure \cite{resnik} with the Best Match
Average strategy \cite{Couto2009} for human and yeast.  The workflow
we followed to predict protein-protein interactions using OPA2Vec is
illustrated in Figure \ref{fig:workflowppi}.
We found that OPA2Vec significantly improves the performance in
predicting interactions between proteins in comparison to both
Resnik's semantic similarity measure and Onto2Vec.

\begin{figure*}[ht]
	\centering
	\begin{subfigure}{.5\textwidth}
		\centering
		\includegraphics[width=\linewidth, trim=4 4 4 4,clip]{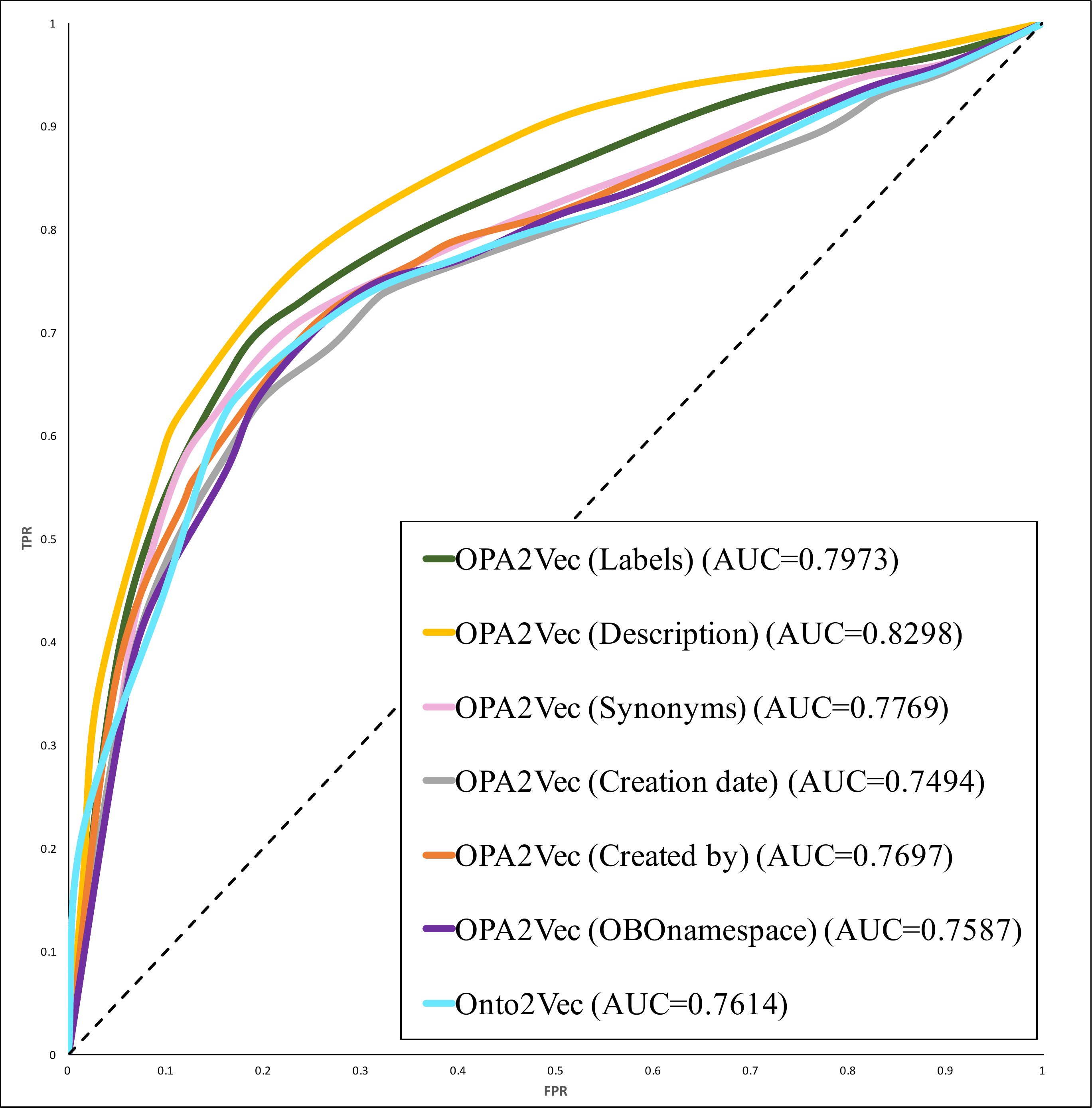}
		\caption{Human}
		\label{fig:sub1}
	\end{subfigure}%
	\begin{subfigure}{.5\textwidth}
		\centering
		\includegraphics[width=\linewidth,trim=4 4 4 4,clip,]{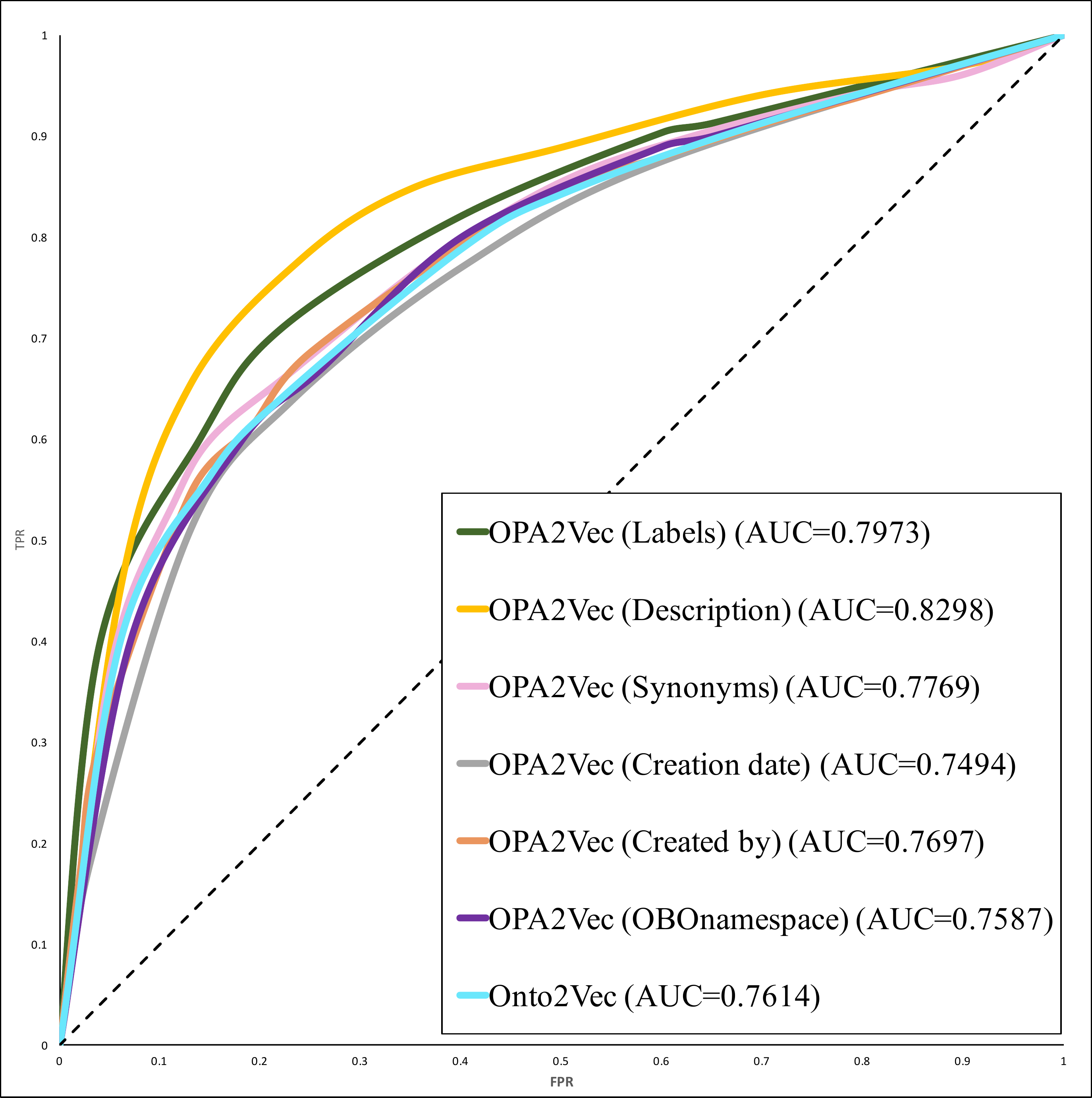}
		\caption{Yeast}
		\label{fig:sub2}
	\end{subfigure}
	\caption{\label{fig:properties}Contribution of each annotation property from the meta-data to the PPI prediction accuracy for human and yeast.}
\end{figure*}

To determine the contribution of each annotation property to the
performance of OPA2Vec, we restricted the inclusion of annotation
properties to each of the following main annotation properties: label
({\tt rdfs:label}), description ({\tt obo:IAO\_0000115}), synonym
({\tt oboInOwl:hasExactSynonym}, {\tt oboInOwl:hasRelated\-Syno\-nym},
{\tt oboInOwl:hasBroadSynonym}, {\tt oboInOwl:hasNarrow\-Synonym}),
created by ({\tt oboInOwl:created\_by}), creation date ({\tt
  oboInOwl:crea\-tion\_date}), and OBO-namespace ({\tt
  oboIn\-Owl:has\-OBONamespace}). Figure \ref{fig:properties} shows the
relative contribution of each of the annotation properties for
prediction of protein-protein interactions. We found that the
inclusion of the natural language descriptions ({\tt
  obo:IAO\_0000115}) and the class labels ({\tt rdfs:label}) results
in the highest improvement of performance, while some annotation
properties such as creation date or the namespace do not improve
prediction.
Interestingly, the {\tt created\_by} annotation property adds some
minor improvement to the performance, likely due to the fact that the
same person would add similar or related classes to the GO, and
therefore proteins with functions created by the same person have
higher probability to interact.

Our analysis shows that annotation properties which describe
biological entities in natural language contribute the most to the
performance improvements of OPA2Vec over Onto2Vec. In particular the
label and description, synonyms and creator ({\tt
  oboInOwl:created\_by}) properties result in better, more predictive
feature vector representations. Therefore, we limited our analysis to
the labels, natural language descriptions, synonyms, and creator name
from the ontology meta-data in further analysis.

We previously found that supervised training can significantly improve
the predictive performance when comparing these vector representations
as it has the potential to ``learn'' custom, task-specific similarity
measures \cite{smaili2018onto2vec}. Therefore, we followed a similar
strategy here and trained a deep neural network (see Section 4.5) to
predict whether two proteins interact given two protein vector
representations as inputs. We found that this supervised approach
further improves the performance of OPA2Vec (Figure
\ref{fig:ppiprediction}).


\subsection{Evaluating performance in predicting gene-disease associations}

As a second use case to evaluate OPA2Vec and demonstrate its utility,
we applied our approach on the PhenomeNET ontology
\cite{rodriguez2017integrating} (see Section 4.1). PhenomeNET is a
system for prioritizing candidate disease genes based on the phenotype
similarity \cite{hoehndorf2011phenomenet} between a disease and a
database of genotype--phenotype associations. Phenotypes refer here to
concrete developmental, morphological, physiological, or behavioral
abnormalities observed in an organism, such as signs and symptoms
which make up a disease \cite{Gkoutos2005, pato-paper}.  The main
advantage of PhenomeNET is that it includes the PhenomeNET ontology
which integrates several species-specific phenotype ontologies; it can
therefore be used to compare, for example, phenotypes observed in
mouse models and phenotypes associated with human disease
\cite{Hoehndorf2013orphanet}.  We used the PhenomeNET ontology and
added mouse genes and human diseases to the knowledge base as
instances; we then associated each instance with a set of phenotypes.
We used the phenotypes associated with unconditional, single gene
knockouts (i.e., complete loss of function mutations) available from
the MGI database \cite{Blake2017} and associated them with their
phenotypes, and we used the disease-to-phenotype file from the HPO
database \cite{hpo} to associate diseases from the Online Mendelian
Inheritance in Men (OMIM) \cite{Amberger2011} database to their
phenotypes. In total, our knowledge base consists of 18,920 genes and
7,154 OMIM diseases.

We applied our OPA2Vec algorithm to the combined knowledge base to
generate vector representations of genes and diseases. We included
only labels, descriptions, synonyms and creators ({\tt created\_by})
as annotation properties as we found them to contribute most to the
performance of OPA2Vec.
The corpus generated by OPA2Vec therefore consists of the set of
asserted and inferred axioms from the PhenomeNET ontology, the set of
annotation axioms involving labels, descriptions, synonyms and
creators, and the gene and disease phenotype annotations.

\begin{figure*}
	\centering
	\begin{subfigure}{.5\textwidth}
		\centering
		\includegraphics[width=\linewidth, trim=4 4 4 4,clip]{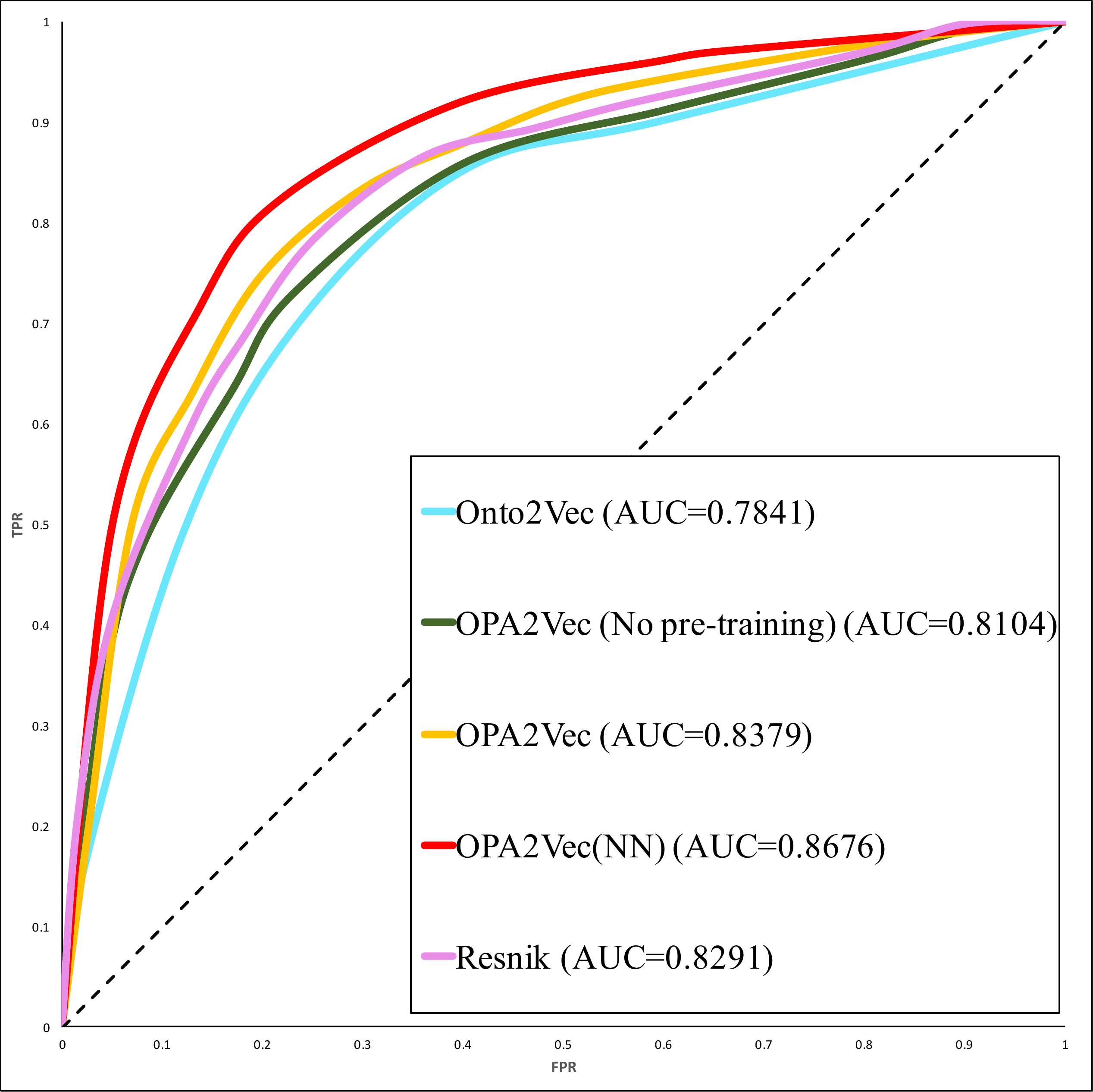}
		\caption{Human}
		\label{fig:sub1}
	\end{subfigure}%
	\begin{subfigure}{.5\textwidth}
		\centering
		\includegraphics[width=\linewidth,trim=4 4 4 4,clip,]{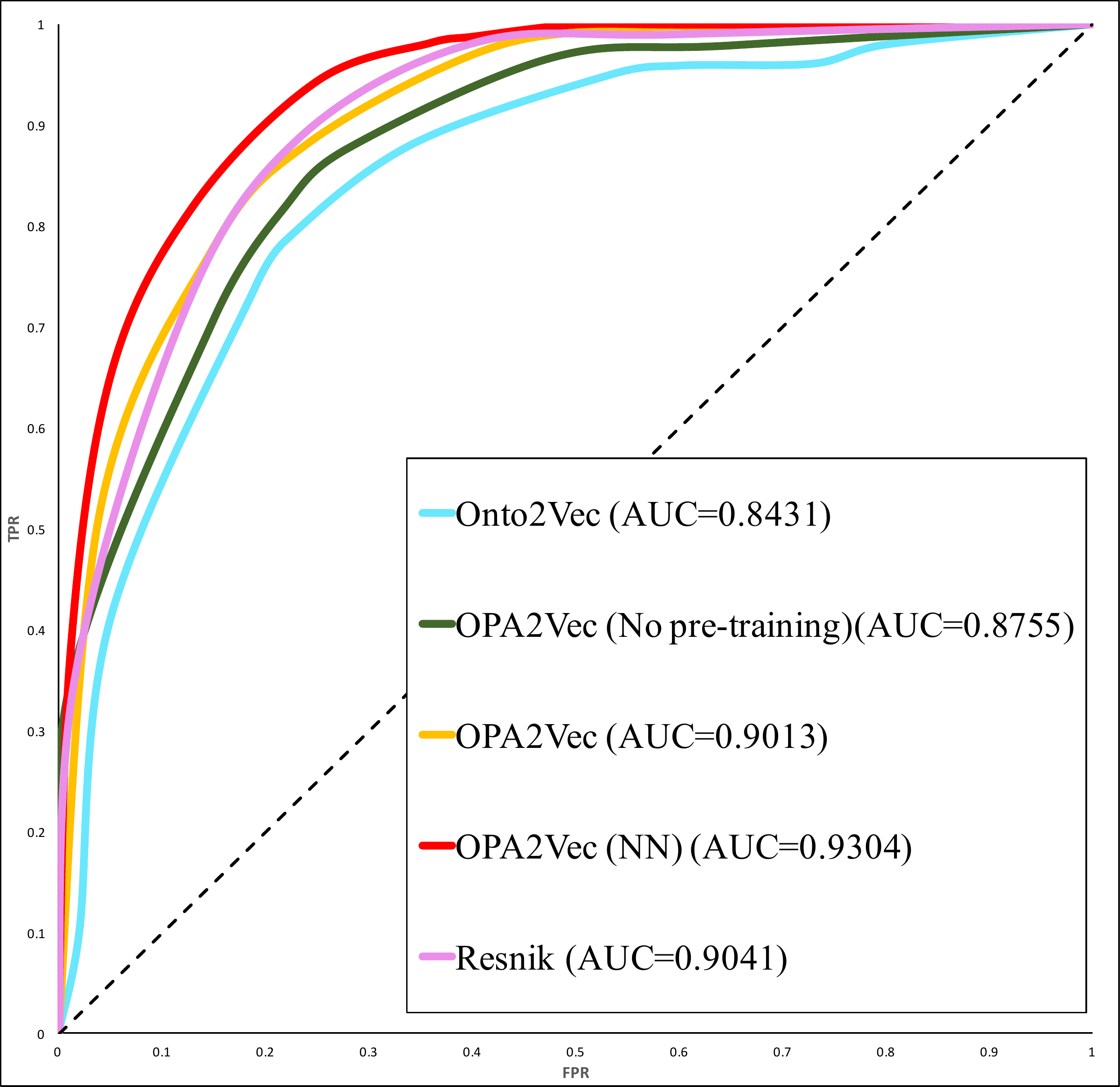}
		\caption{Mouse}
		\label{fig:sub2}
	\end{subfigure}
	\caption{\label{fig:genedisease}ROC curves and AUC values for gene-disease association prediction for different methods for human and mouse.}
\end{figure*}

We then computed the pairwise cosine similarity between gene vectors
and disease vectors, and we trained a neural network in a supervised
manner to predict gene-disease associations. We evaluated our results
using two datasets of gene-disease associations provided by the MGI
database, one containing human disease genes and another containing
mouse models of human diseases.  Figure \ref{fig:genedisease} shows the
ROC curves and AUC values for gene-disease prediction performance of
each approach on the human disease genes and mouse models.  We compared
the obtained results to Resnik similarity and Onto2Vec, and found
that OPA2Vec outperforms both Resnik similarity and Onto2Vec in both
evaluation sets.

\section{Conclusion}

We developed the OPA2Vec method to produce vector representations for
biological entities in ontologies based on the formal logical content
in ontologies combined with the meta-data and natural language
descriptions of entities in ontologies.  We applied OPA2Vec to two
ontologies, the GO and PhenomeNET, and we demonstrated that OPA2Vec
can significantly improve predictive performance in applications that
rely on the computation of semantic similarity.  We also evaluated the
individual contributions of each ontology annotation property to the
performance of OPA2Vec-generated vectors. Our results illustrate that
the annotation properties that describe details about an ontology
concept in natural language, in particular the labels and
descriptions, contribute most to the feature vectors. We could show
that transfer learning, i.e., assigning ``meaning'' to words by
pre-training a Word2Vec model on a large corpus of biomedical
literature abstracts, could further significantly improve OPA2Vec
performance in our two applications (prediction of protein-protein
interactions and prediction of gene-disease associations).

OPA2Vec can comprehensively encode for information in ontologies. Our
method is also based on accepted standards for encoding ontologies, in
particular the Web Ontology Language (OWL), and has the potential to
include or exclude any kind of annotation property in the generation
of its features. OPA2Vec also exploits major developments in the
biomedical ontologies community: the use of ontologies as community
standards, and inclusion of both human- and machine-readable
information in ontologies as standard requirements for publishing
ontologies \cite{Smith2007, Matentzoglu2018}.
We therefore believe that OPA2Vec has the potential to become a highly
useful, standard analysis tool in the biomedical domain, supporting
any application in which ontologies are being used.


\begin{methods}
\section{Methods}
\subsection{Ontology and annotation resources}
We downloaded the Gene Ontology (GO) \cite{GO} in OWL format from
\url{http://www.geneontology.org/ontology/} on September 13, 2017. We
downloaded the GO protein annotations from the UniProt-GOA website
(\url{http://www.ebi.ac.uk/GOA}) on September 26, 2017. We removed all
annotations with evidence code {\tt IEA} as well as {\tt ND}. For
validation, we used the STRING database \cite{string} to obtain
protein-protein interaction (PPI) data for human ({\em Homo sapiens})
and yeast ({\em Saccharomyces cerevisiae}), downloaded on September
16, 2017. The yeast PPI network contains 2,007,135 interactions with
6,392 unique proteins, while the human PPI network contains 11,353,057
interactions  for 19,577 unique proteins.

We downloaded the PhenomeNET ontology
\cite{hoehndorf2011phenomenet,rodriguez2017integrating} in owl format
from the AberOWL repository \url{http://aber-owl.net}
\cite{hoehndorf2015aber} on February 21, 2018. We downloaded the mouse
phenotype annotations from the Mouse Genome Informatics (MGI) database
\url{http://www.informatics.jax.org/} \cite{smith2015expanding} on
February 21, 2018. We obtained a total of 302,013 unique mouse
phenotype annotations. We obtained the disease to human phenotype annotations
on February 21, 2018 from the Human Phenotype Ontology (HPO) database
\url{http://human-phenotype-ontology.github.io/}
\cite{robinson2008human}. We downloaded only the OMIM disease to human phenotype annotations which resulted in a total of 78,208 unique
disease-phenotype associations. For gene-disease association
prediction validation, we used the {\tt MGI\_DO.rpt} file from the MGI
database. This file contains 9,506 mouse gene-OMIM disease
associations and 13,854 human gene-OMIM disease associations. To map
mouse genes to human genes we used the {\tt HMD\_HumanPhenotype.rpt}
file from the MGI database.

To process our ontologies (GO and PhenomeNET), we used the OWL API
4.2.6.\cite{horridge2011owl} and the Elk OWL reasoner
\cite{kazakov2012elk}.

\subsection{PubMed}
We retrieved the entire collection of article abstracts in the MEDLINE
format from the PubMed database
\url{https://www.ncbi.nlm.nih.gov/pubmed/} on February 6, 2018. The
total number of abstracts collected is 28,189,045. For each abstract,
we removed the meta-data (publication date, journal,
authors, PMID, etc.), and only kept the title of the article and the
text of the abstract.

\subsection{Word2Vec}
We used the ontologies, the entity annotations as well as the PubMed
abstracts as the text corpora. To process this text data we used Word2Vec
\cite{word2vec1,word2vec2}. Word2Vec is a machine learning model based
on neural networks that can be used to generate vector representations
of words in a text. Word2Vec is optimized in such a way that the
vector representations of words with a similar context tend to be
similar. Word2Vec is available in two different models: the continuous
bag of word (CBOW) model and the skip-gram model. In this work, we
opted for the skip-gram model which has the advantage over the CBOW
model of creating better quality vector representations of words which
are infrequent in the corpus. This advantage is quite useful in our
case since the biological entities we want to get representations for
do not necessarily occur frequently in our text corpora.  In this
work, we pre-trained the Word2Vec model on the set of PubMed abstracts
and save the obtained model which we eventually retrained on the
ontology studied (the GO ontology and the PhenomeNET ontology). We
used gridsearch to optimize the set of parameters of the skip-gram
model used in this work. We used the same parameters to train Word2Vec
on the PubMed data set and the ontologies data set, except for the
$min\_count$ which has value 25 for the PubMed model, but changed to 1 before training on the ontology corpus. The parameters we
chose are shown in Table {\ref{Tab:word2vec}}.

\begin{table}[!h]
	{\begin{tabular}{@{}lp{5cm}ll@{}}\toprule Parameter &
			Definition & Default value\\\midrule
			\vbox{\hbox{\strut $sg$}}& \vbox{\hbox{\strut Choice of training algorithm}\hbox{\strut $sg$= 1 skip-gram}\hbox{\strut $sg$= 0 CBOW}} & 1 \\
			$size$ & Dimension of the obtained vectors & 200 \\
			$min$\_$count$ &Words with frequency lower than this value will be ignored  & 1 \\ 
			$ window$&Maximum distance between the current and the predicted word & 5\\
			$iter$&Number of iterations&5\\
			$negative$ & Whether negative sampling will be used and how many ``noise words'' would be drawn& 5\\\botrule
	\end{tabular}}{}
	\caption{Parameters used for training the Word2Vec model.\label{Tab:word2vec}} 
      \end{table}
      
\subsection{Similarity}

\subsubsection{Cosine Similarity}
To calculate similarity between the vectors produced by Word2Vec, we
used the cosine similarity which measures the cosine angle between the
two vectors. Cosine similarity $cos_{sim}$ between two vectors $A$ and
$B$ is calculated as 
\begin{equation}
cos_{sim}(A,B)=\frac{A\cdot B}{ ||A||  ||B||},\label{eq:02}
\end{equation}
where $A \cdot B$ is the dot product of $A$ and  $B$.

\subsubsection{Semantic similarity}
Resnik semantic similarity measure \cite{resnik} is one of the most
widely used semantic similarity measures for ontologies. This measure
is based on the notion of information content which quantifies the
specificity of a given concept (term) in the ontology. The information
content of a concept $c$ is commonly defined as the negative log
likelihood, $- log\,p(c)$, where $p(c)$ is the probability of
encountering an instance of concept $c$. Defining information content
in this way makes intuitive sense since as probability increases, the
more abstract a concept becomes and therefore the lower its
information content. Given this definition of information content,
Resnik similarity is formally defined as:
\begin{equation}
sim(c_1,c_2)=-log\,p(c_{MICA}),\label{eq:03}
\end{equation}
where $c_{MICA}$ is the most informative common ancestor of $c_1$ and
$c_2$ in the ontology hierarchy, defined as the common ancestor with
the highest information content value. Resnik similarity measure does
not only have the advantage of being conceptually simple, but it also
overcomes the limitation of assuming that all relations represent
uniform distances, since in real ontologies, the value of one edge may
vary.

Biological entities can have several concept annotations within an
ontology. For instance, as a protein can be involved in different
biological processes and can carry several molecular functions, it can
be annotated by more than one GO terms. Therefore, to calculate
semantic similarity between a pair of proteins, or a pair of any
biological entities, it is necessary to properly aggregate the
similarity between the concepts that they are respectively annotated
with. One possible way to achieve this would be to calculate the Best
Match Average (BMA) which estimates the average similarity between the
best matching terms of two concepts \citep{bma}. For two biological
entities $e_1$ and $e_2$, the BMA is given by:
\begin{equation}
\resizebox{.9\hsize}{!}{$BMA(e_1,e_2)=\frac{1}{2}\bigg(\frac{1}{n}\sum_{c_1\in S_1}\max_{c_2\in S_2}sim(c_1,c_2)+\frac{1}{m}\sum_{c_2\in S_2}\max_{c_1\in S_1}sim(c_1,c_2)\bigg)$},\label{eq:04}
\end{equation}
where $S_1$ is the set of ontology concepts that $e_1$ is annotated
with, $S_2$ is the set of concepts that $e_2$ is annotated with, and
$sim(c_1,c_2)$ is the similarity value between concept $c_1$ and
concept $c_2$, which could have been calculated using the Resnik
similarity or any other semantic similarity measure.

\subsection{Supervised Learning}
To improve our PPI prediction and gene-disease association prediction
performance, we used a neural network algorithm to train our prediction
model. For our PPI prediction, we used  1,015 proteins from the yeast
data set for training and 677 randomly selected proteins were used for
testing while 2,263 proteins from the human data set were used for
training and 1,509 for testing. The positive pairs were all those
reported in the STRING database, while the negative pairs were randomly
sub-sampled among all the pairs not occurring in STRING, in such a way the cardinality of the positive set and the one of the negative
set are equal for the testing and the training datasets.

For the gene-disease association prediction, 6,710 gene-disease
associations were used for training and 2,876 were used for testing
for the mouse gene-disease association prediction. While for the human
gene-disease association prediction, 9,698 associations were used for
training and 4,196 for testing. The positive gene-disease association
pairs were obtained from the {\tt MGI\_DO.rpt} file; all other
associations were considered to be negative.We chose our neural network to be a feed-forward network with four layers: the
first layer contains 400 input units; the second and third layers are
hidden layers which contain 800 and 200 neurons, respectively; and the
fourth layer contains one output neuron. We optimized parameters using
a limited manual search based on best practice guidelines
\cite{hunter2012selection}. We optimized the ANN using binary cross
entropy as the loss function.

\subsection{Evaluation using ROC curve and AUC}
To evaluate our PPI and gene-disease prediction, we used the ROC
(Receiver Operating Characteristic) curve which  is a widely used
evaluation method to assess the performance of prediction and
classification models. It plots the true-positive rate (TPR or
sensitivity) defined as $TPR=\frac{TP}{TP+FN}$ against the
false-positive rate (FPR or $1-$specificity) defined as
$FPR=\frac{FP}{FP+TN}$, where $TP$ is the number of true positives,
$FP$ is the number of false positives and $TN$is the number of true
negatives. Ideally, a perfect classification model would have a ROC
curve that connects the points $(0,0)$, $(0,1)$ and $(1,1)$
\cite{roc}. Generally, the closer the ROC curve bends towards this
``perfect curve'' the better the model is. In the context of this work,
the ROC curve is used to evaluate PPI prediction of our method as well
as competing methods. In this context, the $TP$ value is the number of
protein pairs occurring in STRING regardless of their STRING
confidence score which have been predicted as interacting. The
$FP$ value is the number of protein pairs which have been predicted as
interacting but do not appear in the STRING network and finally the
$TN$ is the number of protein pairs predicted as non-interacting 
which do not occur in the STRING database.

In most cases, ROC curves of different methods would most probably
overlap which makes the visual test of the ROC curves insufficient to
make a formal comparison between different methods \citep{roc}. Thus there is a need for a quantitative measure that
summarizes the meaning of a ROC curve and allows more formal comparison between different methods. The most popular such measures is the
area under the ROC curve (AUC) which is the integration of the ROC
curve over the entire FPR axis \citep{roc}. In this work, the AUC has
also been used along with the ROC curve to evaluate the PPI prediction
performance.
\end{methods}



%
%

\vspace*{-10pt}

\section*{Funding}

The research reported in this publication was supported by the King Abdullah
University of Science and Technology (KAUST) Office of Sponsored Research
(OSR) under Award No. FCC/1/1976-04, FCC/1/1976-06, URF/1/2602-01, URF/1/3007-01, URF/1/3412-01, URF/1/3450-01 and URF/1/3454-01.\vspace*{-12pt}

\bibliographystyle{natbib}

\bibliography{document}

\end{document}